\pdfoutput=1

\documentclass[11pt]{svmult}

\usepackage{emnlp2021}

\usepackage{times}
\usepackage{latexsym}

\usepackage[T1]{fontenc}

\usepackage[utf8]{inputenc}

\usepackage{microtype}

\usepackage{amsmath}
\usepackage{xcolor}
\usepackage{colortbl}
\usepackage{soul}
\usepackage{booktabs}
\usepackage{tikz}
\usetikzlibrary{arrows,shapes,snakes,automata,backgrounds,petri, positioning}

\usepackage{graphicx}        
\usepackage{multicol}        
\usepackage[bottom]{footmisc}

\newcommand{\coref}{coref}
\newcommand{\ellipsis}{ellipsis}
\newcommand{\inc}{inc}
\newcommand{\pronoun}{pronoun}

\setcitestyle{square,numbers}

\title{Active Learning and Multi-label Classification for Ellipsis and Coreference Detection in Conversational Question-Answering}

\author{Quentin Brabant, Lina Maria Rojas-Barahona \\
  Orange Labs, Lannion, France \\
  \texttt{quentin.brabant@orange.com}\\
  \texttt{linamaria.rojasbarahona@orange.com} \And
  Claire Gardent \\
  CNRS/LORIA, Nancy, France \\
  \texttt{claire.gardent@loria.fr}
}

\begin{document}
\maketitle
\begin{abstract}
    In human conversations, ellipsis and coreference are commonly occurring linguistic phenomena. Although these phenomena are a mean of making human-machine conversations more fluent and natural, only few dialogue corpora contain explicit indications on which turns contain ellipses and/or coreferences. In this paper we address the task of automatically detecting ellipsis and coreferences in conversational question answering. We propose to use a multi-label classifier based on DistilBERT. Multi-label classification and active learning are employed to compensate the limited amount of labeled data. We show that these methods greatly enhance the performance of the classifier for detecting these phenomena on a manually labeled dataset.
\end{abstract}

\section{Introduction}

    Human conversations bear inherent complex linguistic phenomena, such as ellipsis and coreferences.
    An ellipsis is the omission of one or several words in a sentence without disturbing its sense in the context. A coreference is the phenomenon occurring when two expressions of a text refer to the same entity (a typical case is when a pronoun is used).
    A proper use of ellipsis and coreferences can make a sentence more concise and easy to process for humans.
    Thus, providing dialogue systems with the ability to use these phenomena in their responses could yield more fluent and natural interactions.

    
    Although ellipsis and coreference have been largely studied in computational linguistics (e.g. \cite{lee-etal-2017-end}), little work has been done towards their study and incorporation in dialogue systems \cite{ultes_pydial_2017,lee_convlab_2019}.
    One difficulty for
    training a model
    to generate dialogue turns that contain ellipses and coreferences is the lack of labeled data: most dialogue corpora contain no explicit indication on which turns contain one of these phenomena and which do not.
    We aim at bridging this gap, by automatically detecting ellipsis and coreferences in dialogue corpora.
    In this paper, we focus on conversational question answering corpora.
    
    Conversational question answering studies the integration of \textit{question answering} (QA) in \textit{dialogue systems} (DS) \cite{reddy_coqa_2019,choi2018quac}. In contrast to task-oriented dialogues, conversational question answering gathers a sequence of coherent questions and answers about almost any topic stored in on-line resources (e.g., Wikipedia or Wikidata), producing open-ended conversations.
    
    We make several contributions to the task of ellipsis and coreference detection in dialogue corpora. We create labeled data by enriching three existing dialogue datasets with annotations indicating whether a turn contains an ellipsis and/or a coreference. As these annotations are incomplete (not every turn can be automatically labelled), we draw on inferential relations between incompleteness, pronominalisation, ellipsis and coreference to both extend (4 labels rather than 2) and complement (specify values for undetermined classes) these annotations.
    We then use these annotated data to train a classifier based on DistilBERT~\cite{sanh_distilbert_2020},
    which assigns to each question in a dialogue two labels indicating whether it contains an ellipsis and/or a coreference.
    We also explore how active learning, multilabel approaches and fine-tuning can be used to train this model.
    
    

\section{Related Work}\label{s:previous_work}
    Most of the work in the literature has focused mainly on the resolution of coreferences for documents~\cite{ng-2010-supervised,lee-etal-2017-end}. The shared task CoNLL 2012 was devoted to the resolution of coreferences for English documents~\cite{pradhan-etal-2012-conll}.
    In~\cite{rojas-barahona-etal-2019-spoken} a conversational question answering system uses a neural end-to-end coreference resolution model trained on documents~\cite{lee-etal-2017-end}.
    
    Recently, however, a few works addressed ellipsis and coreference in dialogues.
    In \cite{gao_interconnected_2019}, a conversational question generation model is proposed; this model is specifically trained to make use of pronominal coreferences.
    \cite{quan_gecor_2019} addresses ellipsis and coreference resolution in dialogues as a variant of question-in-context rewriting.
    This latter task consists in rewriting dialogue turns into pragmatically complete sentences (also called self-contained sentences), i.e., sentences that can be understood outside the context of previous dialogue turns.
    The paper introduces a supervised dataset (see Section \ref{s:datasets}), as well as the GECOR model.
    The task of question-in-context rewriting is also addressed in \cite{elgohary_can_2019}, which introduces the large supervised dataset CANARD.

    Unlike previous work, we formulate ellipsis and coreference detection as a multilabel classification problem.
    For addressing it, we label existing corpora via automated methods and human-in-the-loop approaches (active learning).
    More details are provided in following sections.
    

    
    
    
    
\section{Task: Detection of Ellipsis and Coreferences}\label{s:task}
\label{s:task}

    We are interested in training a classifier for detecting ellipses and coreferences occurring in conversational questions answering corpora.

    A \emph{coreference} occurs when an entity is referred via two or more expressions in the same text (here, the same dialogue). However, we are only interested in detecting a particular kind of coreference. In this paper, we say that a coreference happens in a dialogue turn if and only if (1) it contains an expression referring to an entity already mentioned in a previous dialogue turn and (2) this entity cannot be identified outside of the context of the dialogue. The \emph{resolution} of a coreference consists in replacing the referring expression by an unambiguous reference to the entity.
    
    In linguistics, an \emph{ellipsis} is the omission of one or several words from a clause that preserves the interpretability in context.
    On the contrary, ellipsis (like coreference) does not preserve the interpretability out of context.
    When a dialogue turn is not understandable without its original context (i.e. without the conversation history), we call it \emph{incomplete}.
    In this paper, we assume that any dialogue turn contains an ellipsis if and only if it is still incomplete after coreferences have been resolved.
    It follows from this definition that an incomplete sentence contains either a coreference, an ellipsis, or both. 

    We are mostly interested in detecting ellipsis and coreference in conversational question-answering dialogues.
    Therefore, we treat any dialogue as a sequence of alternating questions and answers that starts with a question and ends with and answer: $(q_1, a_1, q_2, a_2 \dots , q_n, a_n)$.
    In many available conversational question answering datasets questions are natural language sentences produced by humans (e.g. \cite{choi_quac_2018,christmann_look_2019,elgohary_can_2019,quan_gecor_2019,reddy_coqa_2019}), while answers are often given by an automated system, and often not in the form of a sentence.
    For this reason, we focus on ellipsis and coreference detection in questions.
    Moreover, we will sometimes use the term question to refer to dialogue turns that are not question per-say, but are produced by a human interlocutor, as opposed to a dialogue automated system (see Subsection \ref{s:original}, GECOR dataset).
    
    We propose a model whose purpose is to predict whether any given question $q_i$ of a dialogue $(q_1, a_1,  \dots, q_n, a_n)$ contains an ellipsis and/or a coreference;
    since any dialogue turn can normally be understood based on the context of previous turns, our task can be seen as the classification of $q_i$ with the given context $c = (q_1, a_1, \dots , q_{i-1}, a_{i-1})$.
    We thus formulate our task as a $2$-labels classification:
    for a given input question $q_i$ and an input context $c$, output two values $(\coref, \ellipsis) \in \{0,1\}^2$ where $1$ denotes the presence of the phenomenon and $0$ denotes its absence. 
    We call \emph{instance} of our task the couple formed by a question, and its context.
    An instance is \emph{annotated} when it is associated with an annotation of the form $(\coref, \ellipsis)$.

\section{Datasets and annotations}\label{s:datasets}
    In this section we describe the three datasets that we used and how we 
    extracted labeled instances of our task from them.
    We use the following values: 1 for the presence of a phenomenon (positive class), 0 for its absence (negative class). Cases where no label is assigned are denoted by the value -1. Note that -1 does not denote a class, but only the absence of information about the actual class.
    
    \subsection{Original datasets}\label{s:original}
    
    {\bf ConvQuestions}\footnote{\url{https://convex.mpi-inf.mpg.de/}} \cite{christmann_look_2019}
        contains question-answering dialogues. Each dialogue is centered on a ``topic'' entity belonging to one of 5 domains: books, soccer, music, TV series.
        The train/dev/test sets of the dataset contain 33600/11200/11200 questions, respectively.
        The dataset does not originally contain annotations concerning coreference and ellipsis.

    \noindent
    \textbf{GECOR dataset}\footnote{\url{https://github.com/terryqj0107/GECOR}} \cite{quan_gecor_2019}
        is based on the CamRest676 dataset \cite{wen_network-based_2017} and
        contains 676 task-oriented dialogues where an automated system assists a user in finding a restaurant.
        Although not all user turns are questions per say,
        we use them as such during training,
        because they are natural sentences produced by a human.
        Each user turn is associated to up to three variants:
        a completed version (without coreference nor ellipsis),
        a version using coreference, and
        a version using ellipsis.
        Note that some of these variants can be missing or identical to the original; for example, if the original sentence already contains an ellipsis (resp. a coreference), then it can be equal to the elliptical (resp. coreferential) variant. If the original sentence contains no ellipsis (resp. coreference) and there is no satisfying way to introduce one, the elliptical (resp. coreferential) variant is not provided.
        The dataset contains 2744 questions in total (without counting the variants).
        
    \noindent
    \textbf{CANARD}\footnote{\url{https://sites.google.com/view/qanta/projects/canard}} \cite{elgohary_can_2019}
        is based on QuAC \cite{choi_quac_2018}, a question-answering dialogue dataset where each conversation is based on a section of a Wikipedia article.
        Train/dev/test sets repectively contain 31,538/3,418/5,571 questions.
        Each of these questions is provided with a pragmatically complete variant.
        
    \subsection{Instance extraction and labelling}
        Below, we describe how we processed each datasets in order to obtain instances of our task.
        
        \noindent
        {\bf ConvQuestions.}
        We decided to annotate the dialogues from the dataset.
        However, many dialogues of ConvQuestions are centered on the same entity; those dialogues tend to be similar to each others, as they often have questions in common.
        In order to maximize the benefits of manual annotations,
        we created subsets of the original data containing exactly one dialogue per topic entity.
        This resulted in train/dev/test sets containing respectively 905/330/335 questions in total.
        Based on these new sets,
        we created an instance of our task for each question (except the first one) of each dialogue.
        Some of these dialogues where manually annotated with $(\coref, \ellipsis)$ values.
        We obtained train/dev/test of 247/329/331 annotated instances.
        Table \ref{tab:convq-example} provides an example of dialogue and of the corresponding annotated instances.
        
        \begin{table*}
    \setlength{\tabcolsep}{4pt}
    \centering
    Piece of dialogue from ConvQuestions\\
    \begin{tabular}{rl}
        \toprule
        $q_1$ & Who created The Orville? \\
        $a_1$ & Seth MacFarlane \\ \midrule
        $q_2$ & What network airs it? \\
        $a_2$ & Fox Broadcasting Company \\ \midrule
        $q_3$ & How long does an episode run? \\
        $a_3$ & 44 minute \\ \midrule
        $q_4$ & What was the airdate of the first episode? \\
        $a_4$ & 10 September 2017 \\
        \bottomrule
    \end{tabular}
    ~\\
    ~\\
    Corresponding instances of the task:\\
    \setlength{\tabcolsep}{4pt}
    \begin{tabular}{rc||cc||cccc}
        \toprule
        Context & Question & Coref & Ellipsis & \cellcolor{gray!50}Coref & \cellcolor{gray!50}Ellipsis & \cellcolor{gray!50}Incomp. & \cellcolor{gray!50}Pronoun\\ \midrule
        $(q_1, a_1)$ & $q_2$ & 1 & 0 & 1 & 0 & 1 & 1 \\
        $(q_1, a_1, q_2, a_2)$ & $q_3$ & 0 & 1 & 0 & 1 & 1 & 0 \\
        $(q_1, a_1, q_2, a_2, q_3, a_3)$ & $q_4$ & 0 & 1 & 0 & 1 & 1 & 0 \\
        \bottomrule
    \end{tabular}
    \caption{Example of dialogue from ConvQuestions and the corresponding instances of the task. Columns with gray headers show the result of label filling (see Subsection \ref{s:completing}).}
    \label{tab:convq-example}
\end{table*}
        
        \noindent
        {\bf GECOR.}
        We create instances as follows.
        For each dialogue $(q_1, a_1, \dots, q_n, a_n)$ in the GECOR dataset, each $i \in \{2, \dots, n\}$, and each variant $q'_i \in \{q_i(e), q_i(r), q_i(c)\}$ of the question $q_i$:
        if $q'_i$ is not empty, then we create the instance $((q_1, a_1, \dots, a_{i-1}), q'_i)$ and annotate it with $(\coref, \ellipsis)$ values.
        Those values can sometimes be deduced by using the following rules:
        \begin{itemize}
            \item $q_i(e)$ contains an ellipsis;
            \item $q_i(r)$ contains a coreference;
            \item $q_i(c)$ contains no ellipsis nor coreference;
            \item if $q_i(e) = q_i(r)$ we infer that both $q_i(e)$ and $q_i(r)$ contain an ellipsis and a coreference;
            \item if $q_i(e)$ is empty, we infer that $q_i$ contains no ellipsis and thus $q_i(r)$ neither;
            \item if $q_i(r)$ is empty, we infer that $q_i$ contains no coreference and thus $q_i(e)$ neither.
        \end{itemize}
        These rules are not sufficient to deduce ellipsis and coreference label values in all cases; see for example Table \ref{tab:gecor-example}.
        By default, the value -1 is assigned.
        
        \begin{table*}
    \setlength{\tabcolsep}{4pt}
    \centering
    Piece of dialogue from GECOR dataset:
    \begin{tabular}{rl}
        \toprule
        $q_1$ & I am looking for a restaurant that is moderately priced and serves Cantonese food. \\
        $q_1(e)$ & - \\
        $q_1(r)$ & - \\
        $q_1(c)$ & I am looking for a restaurant that is moderately priced and serves Cantonese food. \\ \midrule
        $a_1$ & There are no restaurants that serve Cantonese food in the moderate price range. \\ \midrule
        $q_2$ & How about chinese type of food? \\
        $q_2(e)$ & How about chinese type of food? \\
        $q_2(r)$ & - \\
        $q_2(c)$ & How about moderately priced chinese type of food? \\ \midrule
        $a_2$ & The Golden wok 
        [...]. Would you like their location? \\ \midrule
        $q_3$ & Their phone number please. \\
        $q_3(e)$ & phone number please. \\
        $q_3(r)$ & Their phone number please. \\
        $q_3(c)$ & I would like the phone number of the Golden wok please. \\ \bottomrule
    \end{tabular}
    ~\\
    ~\\
    Corresponding instances of the task:\\
    \setlength{\tabcolsep}{6pt}
    \begin{tabular}{rc||cc||cccc}
        \toprule
        Context & Question & Coref & Ellipsis & \cellcolor{gray!50}Coref & \cellcolor{gray!50}Ellipsis & \cellcolor{gray!50}Incomp. & \cellcolor{gray!50}Pronoun\\ \midrule
        $(q_1, a_1)$ & $q_2(e)$ & 0 & 1 & 0 & 1 & 1 & 0 \\
        $(q_1, a_1)$ & $q_2(c)$ & 0 & 0 & 0 & 0 & 0 & 0 \\
        $(q_1, a_1, q_2, a_2)$ & $q_3(r)$ & 1 & -1 & 1 & -1 & 1 & 1 \\
        $(q_1, a_1, q_2, a_2)$ & $q_3(e)$ & -1 & 1 & -1 & 1 & 1 & 0 \\
        $(q_1, a_1, q_2, a_2)$ & $q_3(c)$ & 0 & 0 & 0 & 0 & 0 & 0 \\
        \bottomrule
    \end{tabular}
    \caption{Example of dialogue from the GECOR dataset. Columns with gray headers show the result of label filling (see Subsection \ref{s:completing}).}
    \label{tab:gecor-example}
\end{table*}
        
        \noindent
        {\bf CANARD.}
        Instances were extracted similarly as from the GECOR dataset. The two main differences are: for each created dialogue, two variants (original and complete) of the last question are used.
        When the complete variant is used, we assign $0$ to both $\coref$ and $\ellipsis$; otherwise, we assign $-1$.
        An example is given in Table \ref{tab:canard-example}.

        \begin{table*}
    \setlength{\tabcolsep}{4pt}
    \centering
    Piece of dialogue from CANARD (answers $a_1$ and $a_2$ are not shown):\\
    \begin{tabular}{rl}
        \toprule
        $q_1$ & What is On the Sunday of Life? \\
        $q_1(c)$ & What is On the Sunday of Life? \\ \midrule
        $q_2$ & Did it do well? \\
        $q_2(c)$ & Did Porcupine Tree, On the Sunday of Life do well? \\ \midrule
        $q_3$ & Was it rereleaesd? \\
        $q_3(c)$ & Was Porcupine Tree, On the Sunday of Life rereleaesd? \\
        \bottomrule
    \end{tabular}
    ~\\
    ~\\
    Corresponding instances of the task:\\
    \setlength{\tabcolsep}{6pt}
    \begin{tabular}{rc||cc||cccc}
        \toprule
        Context & Question & Coref & Ellipsis & \cellcolor{gray!50}Coref & \cellcolor{gray!50}Ellipsis & \cellcolor{gray!50}Incomp. & \cellcolor{gray!50}Pronoun\\ \midrule
        $(q_1, a_1)$ & $q_2$ & -1 & -1 & 1 & -1 & 1 & 1 \\
        $(q_1, a_1)$ & $q_2(c)$ & 0 & 0  & 0 & 0 & 0 & 0 \\
        $(q_1, a_1, q_2, a_2)$ & $q_3$ & -1 & -1  & 1 & -1 & 1 & 1 \\
        $(q_1, a_1, q_2, a_2)$ & $q_3(c)$ & 0 & 0 & 0 & 0 & 0 & 0 \\
        \bottomrule
    \end{tabular}
    \caption{Example of dialogue from CANARD and the corresponding instances of the task. Columns with gray headers show the result of label filling (see Subsection \ref{s:completing}).}
    \label{tab:canard-example}
\end{table*}
    
    \subsection{Extending and completing annotations}\label{s:completing}
        At this point many labels are missing in the instances of the task.
        In particular, instances from CANARD do not contain any positive label.
        We addressed this issue via two approaches: multilabel learning and label filling.

        Multilabel classification can be seen as a particular case of multitask learning, since a single model is trained on several binary classification tasks. One justification for using this approach (instead of one model per classification) is that related classifications rely on similar sets of features, and thus training on one classification updates the model parameters in a way that is beneficial to the others.

        We already introduced our task as a $2$-labels (coreference and ellipsis) classification task.
        In this subsection we extend it to a $4$-labels classification task where labels are: coreference, ellipsis, incompleteness, and pronoun detection.
        Formally, it means that annotations of the form $(\coref, \ellipsis)$ are replaced by annotations of the form $(\coref, \ellipsis, \inc, \pronoun)$.
        We used automatic pronoun detection to provide a $0$ or $1$ value to $\pronoun$ in all questions.
        By default, the value of $\inc$ is set to -1, except for instances from CANARD where the value is known.
        
        We then replace some of the $-1$ values by taking advantage of the logical dependencies between labels: a pronoun always indicates a coreference; incompleteness is either due to a coreference or an ellipsis; coreferences and ellipses always cause incompleteness. We therefore applied the following rules to each instance, in order:
        \begin{enumerate}
            \item if $\pronoun = 1$ then $\coref \gets 1$,
            \item if $\coref = 1$ or $\ellipsis = 1$ then $\inc \gets 1$,
            \item if $\coref = 0$ and $\ellipsis = 0$ then $\inc \gets 0$,
            \item if $\inc = 0$ then $\coref  \gets 0$ and $\ellipsis  \gets 0$.
        \end{enumerate}
        Remark that in some cases these rule are not sufficient to get rid of all unknown values.
        Such cases can be found in the examples of Tables \ref{tab:gecor-example} and \ref{tab:canard-example}.
        
        The summary of the obtained data is given in Table \ref{tab:data-summary}.
        
        \begin{table*}
            \centering
            \setlength{\tabcolsep}{5pt}
            \begin{tabular}{rcccccccccc}
                \toprule
                &\# instances  & \multicolumn{2}{c}{Coreference} & \multicolumn{2}{c}{Ellipsis} & \multicolumn{2}{c}{Incomp.} & \multicolumn{2}{c}{Pronoun}  \\
                 \midrule
                ConvQuestions & 249/327/331 & 1/1/1 & .54 & 1/1/1 & .32 & 1/1/1 & .78 & 1/1/1 & .31 \\
                GECOR & 4045/549/522 & .8/.8/.8 & .30 & .8/.8/.8 & .33 & 1/1/1 & .49 & 1/1/1 & .13 \\
                CANARD & 59k/6k/10k & .8/.8/.8 & .36 & .5/.5/.5 & 0 & 1/1/1 & .50 & 1/1/1 & .29 \\
                \bottomrule
            \end{tabular}
            \caption{Summary of the labeled data. For each label, values that are given in the form ``$a/b/c$~~ $d$'' represent ratios: $a/b/c$ are the proportions of instances for which the class is known in train/dev/test, respectively; $d$ is the proportion of positive values among instances where the class is known (in the union of train, dev, and test sets).}
            \label{tab:data-summary}
        \end{table*}

\subsection{Evaluation and training}
    We use GECOR and CANARD for training our models, while ConvQuestions is used for evaluation and fine tuning. In this way we can better assess how well the classifier behaves on unseen data, data that is different from the data the model was trained on.
    During training, labels with -1 value are simply ignored (no error is retro-propagated).
    During evaluation, we measure the recall, precision, and F-measure on ellipsis and coreference detection.

\section{Proposed Models}\label{s:model}
    
    We propose a baseline model and several variants: they all share the same architecture but differ on the data and labels they are trained on.
    In this way, we explore how active learning and fine-tuning impact the classifier performance.
    
    \subsection{Baseline and variants}
    
    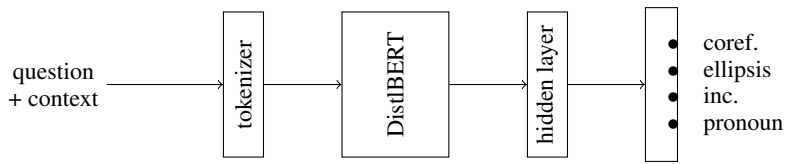
\begin{figure*}
    \centering
    \begin{tikzpicture}
        \tikzstyle{flow}=[->]
        \tikzstyle{block}=[draw,rectangle, align=center, minimum height={width("hidden layer")+10pt}, minimum width=15pt]
        \tikzstyle{ext}=[align=center]
        
        \node[ext] (input) at (-4.5, 0) {question\\ + context};
        \node[block] (tok) at (-2,0) {\rotatebox{90}{tokenizer}};
        \node[block, minimum width=40pt] (db) at (0,0) {\rotatebox{90}{DistlBERT}};
        
        \node[block] (pc) at (2,0) {\rotatebox{90}{hidden layer}};
        \node[block,minimum height=58pt,minimum width=12pt] (out) at (3.5,0) { };
        \node[align=left] (x) at (4.34,0) {$\bullet$~~~ coref.\\ $\bullet$~~~ ellipsis\\ $\bullet$~~~ inc.\\ $\bullet$~~~ pronoun};
        
        \draw[flow] (input) to (tok);
        \draw[flow] (tok) to (db);
        \draw[flow] (db) to (pc);
        \draw[flow] (pc) to (out);
    \end{tikzpicture}
    \caption{Graphical representation of the model.}\label{fig:model}
\end{figure*}
        
    Our model relies on DistilBERT \cite{sanh_distilbert_2020}
    It is composed of several steps: a tokenizer, followed by a DistilBERT model, a fully connected hidden layer, and a 4-unit output layer.
    The context is composed of the 8 last dialogue turns preceding the question; the context and the question are concatenated and given as input to the tokenizer.
    Resulting tokens are then fed to DistilBERT.
    The pooled output of DistilBERT is fully connected to the hidden layer, which is itself fully connected to the output layer.
    Each unit value yields a prediction for one label.

    The baseline is trained on the $4$-labels classification task, on a mixture containing all instances from GECOR and as many from CANARD, to which we refer as CANARD/GECOR.
    Variants from the baseline are obtained by applying one or several of the following modifications:
    \begin{enumerate}
        \item training for 2-labels (ellipsis and coreference) instead of 4,
        \item adding labeled data to CANARD training set, via active learning (see next subsection),
        \item fine-tuning by training on ConvQuestions training set (after training on CANARD/GECOR),
        \item fine-tuning only (no training on CANARD/GECOR).
    \end{enumerate}
    
    \subsection{Active learning}
        Active learning is a human-in-the-loop method that aims at maximizing the performance gains relatively to the number of manual annotations. It is especially interesting when few labeled data are available and only a small fraction of unlabeled data can be manually annotated in reasonable time. We apply several rounds of active learning for labeling (separately) ellipses and coreferences. Each round consists in the following steps:
        \begin{enumerate}
            \item {\it Train and evaluate a model.} We use CANARD/GECOR as a training set. All CANARD instances that have already been manually annotated during previous rounds are included.
            The evaluation is done on ConvQuestions test set.
            \item {\it Run the model on unlabeled data.} The model trained in step 1 associates a prediction $(\coref^*, \ellipsis^*)$ to each instance.
            \item {\it Select a subset of unlabeled data.}
            We select the 50 CANARD dialogues on which the model display the least certainty. Since one dialogue is the source of several instances, 
            we define the certainty of a dialogue as the average certainty of the corresponding instances.
            The certainty of the model (for a given label, on a given instance) is defined as the distance from $0.5$ of the output corresponding to the predicted label value, i.e.: $|\coref^* - 0.5|$ for coreference and $|\ellipsis^* - 0.5|$ for ellipsis.
            \item {\it Manually label the selected subset.} We label the selected dialogues (either for ellipsis or coreference).
            Labeled dialogues are used during training in the next loop.
        \end{enumerate}
        We stop repeating these steps when evaluation score stop increasing.

\section{Experiments}\label{s:experiments}

    The experiments reported in this section have two main purposes: the first is to evaluate the performances of the model on the test set of ConvQuestions; the second is to estimate the usefulness of multilabel and active learning.
    
    
    \subsection{Model variants and hyper parameters}
        We evaluate the following model variants.
        \begin{itemize}
            \item {\it Baseline}.
            The model is trained on the $4$-label classification task on CANARD/ GECOR.
            \item {\it Fine tuning only}.
            The model is trained on the $4$-label classification task on the training set of ConvQuestions.
            \item {\it Baseline + AL.}
            The model is trained on the $4$-label classification task on CANARD/GECOR, but labeled instances of CANARD are added via active learning. Each round of active learning adds 50 instances that are labeled for either coreference or ellipsis. We evaluate several versions of this variant: three versions use instances that were annotated for coreference via, respectively, 1, 2, and 3 rounds of active learning. Three others versions use instances that were annotated for ellipsis via 1, 2, and 3 rounds.
            \item {\it Baseline + all AL.}
            Identical to baseline + AL, but using all annotations produced for coreference and ellipsis (3 rounds for each).
            \item {\it Baseline + all AL + fine tuning}.
            Identical to {\it Baseline + all AL.}, but training on CANARD/GECOR is followed by a fine-tuning step on the training set of ConvQuestions.
            \item {\it 2-label variants}.
            We evaluate three of them. They are respectively identical to { \it baseline}, to {\it baseline + all AL}, and to {\it baseline + all AL + fine tuning}, with the difference that the model is trained on the 2-labels classification task.
        \end{itemize}
        
        We use the ``distilbert-base-uncased'' pretrained HuggingFace model \footnote{\url{https://huggingface.co/distilbert-base-uncased}}.
        The tokenization of the concatenated string is done by HuggingFace's DistilBertTokenizerFast.
        The same hyperparameters are used in all variants.
        The hidden layer is made of 768 units with ReLU activation function.
        The output layer uses sigmoid activation function.
        For all models, training is done in 10 epochs, with a batch size of size 16, a learning rate of $0.0001$, and a dropout probability of $0.1$.
        During training, weight are updated by retropropagating the Mean Squared Error of each output unit.
        We compensate class imbalance by using class weights, in such way that the cumulative weight of negative and positive classes are equal (for ellipsis and coreference, respectively).

    \subsection{Results}
    
        The results are displayed in Table \ref{tab:results}. Each line corresponds to a variant of the model.
        
        \begin{table*}
            \centering
            \setlength{\tabcolsep}{9pt}
            \begin{tabular}{rlcccccc}
            	\toprule
            	& & \multicolumn{3}{c}{Coreference} & \multicolumn{3}{c}{Ellipsis} \\
            	 & & P & R & F1 & P & R & F1 \\ \midrule
            	1 &  fine tuning only & 81 & 65 & 72 & 51 & 67 & 57 \\ \midrule
            	2 & baseline & \bf 97 & 64 & 77 & 64 & 36 & 46 \\
            	3 & ~~~~~ + AL for ellipsis (1 round) & 92 & 63 & 75 & 71 & 48 & 56 \\
            	4 & ~~~~~ + AL for ellipsis (2 rounds) & 89 & 72 & 80 & 83 & 41 & 55 \\
            	5 & ~~~~~ + AL for ellipsis (3 rounds) & 85 & 79 & 82 & 74 & 48 & 57 \\
            	6 & ~~~~~ + AL for coref. (1 round) & 87 & 84 & 85 & 72 & 46 & 56 \\
            	7 & ~~~~~ + AL for coref. (2 rounds) & 92 & 81 & 86 & 71 & 39 & 50 \\
            	8 & ~~~~~ + AL for coref. (3 rounds) & 95 & 79 & 86 & 67 & 31 & 43 \\
            	9 & ~~~~~ + all AL labels & 94 & 81 & 87 & 84 & 46 & 59 \\
            	10 & ~~~~~ + fine tuning & 94 & \bf 93 & \bf 94 & 83 & \bf 71 & \bf 77 \\ \midrule
            	11 & baseline, 2-labels variant & 89 & 68 & 77 & \bf 100 & 10 & 19 \\
            	12 & ~~~~~ + all AL labels & 91 & 86 & 89 & 88 & 35 & 50 \\
            	13 & ~~~~~ + all AL labels + fine-tuning & 94 & \bf 93 & 93 & 84 & 70 & 76 \\
            	\bottomrule
            \end{tabular}
            \caption{Results of the experiments. Scores are given as percentages.}
            \label{tab:results}
        \end{table*}
        
        Generally, the results show that coreference detection performs better than ellipsis detection.
        Moreover, by looking at lines 2 to 9 in the table, we see that active learning is clearly beneficial; the {\it all AL labels} variant improves F1 scores for coreference and ellipsis detection by $10$ and $13$ points compared to the baseline.
        The same conclusion is drawn when comparing lines 11 and 12.
        The effects of training on 4 labels versus 2 are less clear:
        by comparing lines 2, 9, 10 to lines 11, 12, 13, we see that 4-labels variants perform roughly as well as their 2-labels counterparts on coreference detection.
        For ellipsis detection, they score significantly higher on F1 score when no fine tuning is applied, but the scores are too low to propose a meaningful interpretation.
        Fine tuning increases scores for both ellipsis and coreference detection; however the increase is way larger in the case of ellipsis. In fact, coreference detection arguably performs reasonably well without fine-tuning, contrary to ellipsis detection. A possible explanation is that the kinds of ellipses occurring in one dataset can be different from those occurring in another. In contrast, coreferences cover a narrower set of phenomena.

        In addition to measuring performances, we looked at the output of the model on the test set: we noticed that coreferences due to pronouns use are well recognized, while, many false negatives correspond to cases where an entity is referred to via its type or function, as in: ``To which continent does Germany belong? What size is the country?''.

\section{Conclusion}\label{s:conclusion}
    In this work, we proposed an ellipsis and coreference detection model based on DistilBERT. We have shown that it is possible to obtain reasonable performance for coreference detection on an unknown conversational QA dataset. Unsurprisingly, our results also show that using only active learning already improves the F1 measure for ellipsis and coreference detection.
    Our model is significantly better at detecting coreferences than ellipses on an unknown dataset.
    Since ellipsis in fact cover a wide range of different phenomena, a possible way of improvement could be to differentiate several kind of ellipses and train the model to recognize each of them.


\bibliographystyle{spbasic}
\bibliography{iwsds}

\end{document}